\documentclass{dataset}
\usepackage{pifont}
\usepackage{amssymb}
\usepackage{authblk}

\usepackage[backend=biber, style=numeric, sorting=none]{biblatex}
\title{Paired Uterine Whole-Slide Images and Pathology Reports for Multimodal Computational Pathology}

\author[1,2,6,\dag,*]{Han Li}
\author[1,6,\dag]{Jingsong Liu}
\author[1,3]{Ayako Ura}
\author[5]{Junlin Hou}
\author[1,6]{Zhengyang Xu}
\author[1,6]{Azar Kazemi}
\author[1,6]{Oskar Thaeter}
\author[1,6,7]{Christian Grashei}
\author[1,6]{Fabian Gülhan}
\author[1,6]{Reza Nasirigerdeh}
\author[2]{Xun Ma}
\author[4]{Rui Yan}
\author[5]{Hao Chen}
\author[4]{S. Kevin Zhou}
\author[2,6]{Nassir Navab}
\author[1]{Carolin Mogler}
\author[1,6,7]{Peter Sch\"uffler}

\affil[1]{Institute of Pathology, Technical University of Munich, Munich, Germany}
\affil[2]{Computer Aided Medical Procedures (CAMP), Technical University of Munich, Munich, Germany}
\affil[3]{Department of Human Pathology, Juntendo University Graduate School of Medicine, Tokyo, Japan}
\affil[4]{School of Biomedical Engineering, University of Science and Technology of China, Hefei, China}
\affil[5]{The Hong Kong University of Science and Technology, Hong Kong, China}
\affil[6]{Munich Center for Machine Learning (MCML), Munich, Germany}
\affil[7]{Munich Data Science Institute (MDSI), Munich, Germany}

\begin{document}

\maketitle

\begingroup
\renewcommand\thefootnote{}
\footnotetext{
\textsuperscript{\dag} These authors contributed equally.
\phantom{\textsuperscript{\dag}}\textsuperscript{*} Corresponding author.
}
\endgroup

\begin{abstract}

Uterine diseases represent an important category of gynecologic pathology and require accurate histopathological assessment for diagnosis and treatment planning. Whole-slide images (WSI) have enabled the digital transformation of pathology workflows and provided new opportunities for artificial intelligence (AI) in computational pathology. In particular, multimodal models that jointly analyze histopathology images and pathology reports have shown promising potential for automated pathology report generation and AI-assisted diagnosis.
However, the development of such systems remains limited by the scarcity of datasets that pair whole-slide images with clinically meaningful pathology reports.
Instead, existing pathology datasets focus on patch- or slide-level annotations of a single endpoint (e.g., disease class), which do not fully capture the rich information in full clinical diagnostic workflow reports.
Here, we introduce TUM-Uteria, a uterine pathology dataset comprising WSIs paired with diagnostic pathology reports at both the case and slide levels, collected from a tertiary medical center. The dataset contains 216 clinical cases, comprising 455 slide-level WSI–report pairs. 
The dataset underwent a structured multi-stage validation procedure involving board-certified pathologists to ensure reliable annotations. TUM-Uteria supports research in computational pathology, including whole-slide image analysis, multimodal learning, and automated pathology report generation.

\end{abstract}

\section{Background \& Summary}

Uterine diseases, including endometrial carcinoma and other gynecologic tumors, represent a significant global health burden \cite{amant2005endometrial}. Histopathological examination remains the gold standard for diagnosing these diseases and for determining tumor type, grade, and stage, which are essential for guiding treatment decisions \cite{aljehani2023importance}. Pathologists typically examine tissue sections under the microscope and produce diagnostic reports describing morphological findings and clinical conclusions \cite{campbell2014semantic}.

With the increasing adoption of digital pathology, whole-slide images (WSIs) have become widely available in clinical practice \cite{hanna2019whole, pantanowitz2011review}. These gigapixel-resolution images enable large-scale computational analysis and have stimulated rapid development in artificial intelligence (AI) methods for pathology image interpretation \cite{aeffner2019introduction, lu2022federated}. Deep learning models have demonstrated promising results in tasks such as tumor detection, tissue classification, and cancer subtyping \cite{campanella2019clinical, bokhorst2023deep}.

More recently, multimodal AI systems that integrate visual and textual clinical information have gained increasing attention \cite{qiao2022multi, li2025multi}. Visual–language models capable of jointly analyzing WSIs and pathology reports may provide new opportunities for automated pathology report generation, diagnostic assistance, and clinical decision support \cite{ahmed2024pathalign, hu2025pathology}.

Despite these advances, the development of multimodal computational pathology systems remains constrained by the limited availability of datasets that pair whole-slide images with corresponding pathology reports \cite{lu2021data, ding2025multimodal}. Most existing pathology datasets focus on patch-level annotations or classification tasks and rarely include slide-level image–text pairs reflecting real clinical documentation \cite{lu2021data, ding2025multimodal}. As a result, the development and evaluation of multimodal pathology models remain challenging \cite{xu2025multimodal}.

To address these limitations, we present \textbf{TUM-Uteria}, a uterine pathology dataset that systematically links WSIs to diagnostic pathology reports at both the case and slide levels. The dataset contains \textbf{216} and \textbf{455 slide-level WSI--report pairs} collected from routine diagnostic workflows at a tertiary academic medical center in Germany.
The original diagnostic reports were written in German and subsequently translated into English. Manual quality control was performed to correct translation errors, typographical errors, and formatting artifacts, ensuring that the released English reports preserve the clinical information contained in the original German reports. 

A key feature of TUM-Uteria is that the cases were collected according to routine clinical practice rather than being selectively enriched for malignant or rare diseases. As a result, the dataset reflects the natural spectrum of uterine pathology encountered in a tertiary medical center, including common benign conditions, inflammatory or non-neoplastic findings, premalignant lesions, and malignant tumors. Case-level diagnostic reports serve as the primary source of clinical information and are further processed into slide-level descriptions to establish fine-grained correspondence between visual and textual data. Given the inherent challenges of this alignment, the dataset is constructed through a structured multi-stage validation pipeline involving domain experts at different stages. This process ensures clinically consistent case-level diagnoses and accurate slide-level associations.

These characteristics make \textbf{TUM-Uteria} a valuable resource for computational pathology research, supporting tasks such as whole-slide image analysis, multimodal learning, and automated pathology report generation.

\section{Methods}
\begin{figure*}[th]
  \centering
\includegraphics[width=1\textwidth]{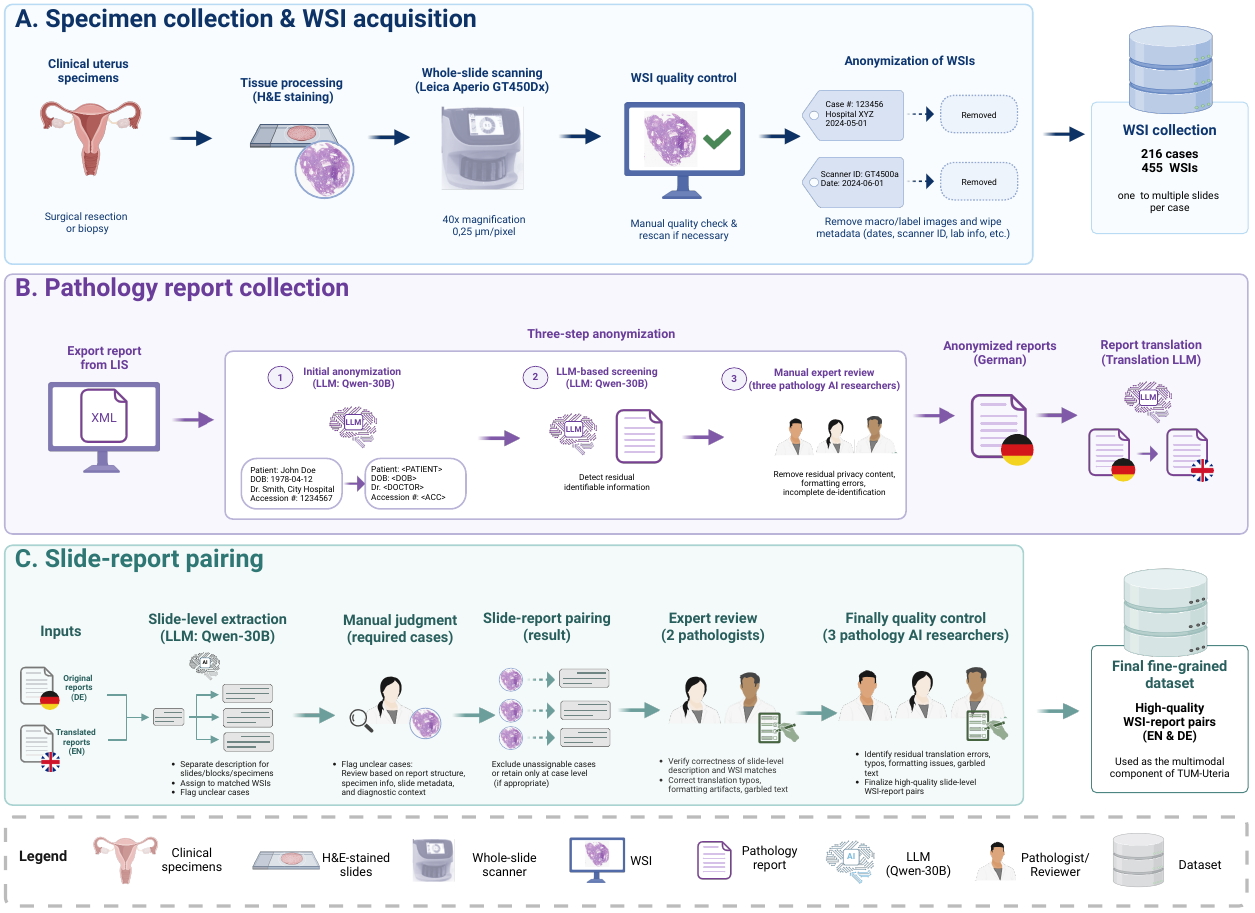}
\caption{
\textbf{Construction workflow of the TUM-Uteria dataset.}
The workflow consists of three main parts:
(A) specimen collection and whole-slide image acquisition, including tissue processing, H\&E staining, slide digitization, image quality control, and WSI anonymization;
(B) pathology report collection and processing, including XML report export, LLM-based report anonymization, manual privacy review, and German-to-English translation;
and
(C) slide--report pairing and quality control, including LLM-assisted slide-level report extraction, manual review of ambiguous WSI--report correspondences, expert validation, and final quality-control assessment to generate the released multimodal WSI--report pairs.
}
 \label{fig1}
\end{figure*}

\subsection{Specimen collection and whole-slide image acquisition}

\textbf{Specimen collection:}
The TUM-Uteria dataset was constructed from uterine pathology cases collected during routine clinical diagnostic workflows at the Institute of Pathology of the Technical University of Munich, Germany. Tissue specimens were obtained through surgical resection or biopsy procedures, including hysterectomy, curettage, biopsy, and other surgical specimens, and were processed according to standard histopathological protocols. The resulting tissue sections were stained with hematoxylin and eosin (H\&E), the standard staining method for visualizing cellular morphology and tissue architecture.

In total, the dataset contains \textbf{216 clinical cases} comprising \textbf{455 whole-slide images}. Each case may include one or multiple slides corresponding to different tissue sections from the same patient. Cases were selected based on the availability of diagnostic pathology reports and digitized H\&E-stained whole-slide images.

\textbf{Whole-slide image acquisition:}
Glass slides were digitized using a clinical whole-slide scanner (Leica Aperio GT450Dx, Leica Biosystems, Buffalo, USA) at 40$\times$ magnification and 0.25 $\mu$m/pixel resolution. All WSIs have been manually controlled for image quality and scanning artifacts. Where applicable, poor-quality image slides have been rescanned.

All WSIs were anonymized by removing the macro and label images to remove case numbers and barcodes, and by wiping out lab-specific information such as scanning date or scanner ID from the WSIs' metadata.

\subsection{Pathology report collection}

Reports for each case were exported from the laboratory information system in an XML format. The original reports contain sections for (i) the clinical question, (ii) the macroscopy, (iii) the microscopic description, and (iv) the critical finding. 
Note that a case may contain one or more tissue blocks and slides, but only one report. To still obtain descriptions for individual slides, we paired slides with reports as described in the next section.

As shown in Fig.\ref{fig1} (B), all reports (in German) were \textbf{anonymized in a three-step approach}. In the first step, a locally deployed, offline LLM (Qwen-30B) was used to identify and replace potentially identifiable information in the reports with placeholder tokens such as '\texttt{xxx}'. This included patient names, doctor's names, hospital names, dates of birth, accession numbers, phone numbers, and other potentially identifying information.

After the initial anonymization, a second LLM-based screening step was performed using Qwen-30B to check whether any residual identifiable information remained in the processed reports. Reports flagged during this step were further revised to remove remaining sensitive information.

Finally, three pathology AI researchers manually reviewed the anonymized reports to identify residual privacy-related content, formatting errors, or incomplete de-identification.

Only reports that passed this three-step anonymization procedure were included in the final dataset. All subsequent report processing, slide-level extraction, translation, and validation steps were conducted using the anonymized reports.

To support a wider use of our dataset, the anonymized German reports were \textbf{translated into English} with offline LLM (Qwen-30B).

\subsubsection{Slide--report pairing}
Although pathology reports often describe findings according to individual slides or tissue blocks, the degree of slide-level structure varies across reporting styles and pathologists. Some reports explicitly indicate which descriptions correspond to specific slides or specimen parts, whereas others provide descriptive findings without clear slide identifiers. Therefore, additional processing was required to convert the original case-level reports into structured slide-level descriptions and to establish reliable associations between each WSI and its corresponding textual description.

As shown in Fig.\ref{fig1} (C), the translated reports (in English) were used together with the original German reports for slide-level report extraction. A locally deployed offline Qwen-30B model was used to assist the extraction of slide-level descriptions from the case-level reports. For reports with explicit slide, block, or specimen identifiers, the model separated the corresponding textual descriptions and assigned them to the matched WSIs. For reports without clear slide-level identifiers, the model flagged the case as requiring manual assessment rather than assigning uncertain WSI--text correspondences automatically.

In total, \textbf{71 cases} required manual judgment to determine the appropriate correspondence between WSIs and textual descriptions. These cases were reviewed based on the report structure, specimen information, slide metadata, and diagnostic context to establish reliable slide-level associations. When a reliable correspondence between an individual WSI and a textual description could be established, the case was included in the slide-level WSI--report pairs. If such correspondence could not be determined with sufficient confidence, no slide-level pair was generated for that WSI, although the corresponding case-level report was retained as part of the case-level dataset.

After the initial slide--report pairing, all cases that did not require manual assignment were further reviewed by two pathology experts to ensure data quality. This review step was designed to confirm that the automatically extracted slide-level descriptions were consistent with the original diagnostic reports and correctly associated with the corresponding WSIs. During this review, translation-related typographical errors, formatting artifacts, and garbled text were also corrected when identified.

After completion of the slide-level pairing and expert review, three pathology AI researchers conducted an additional manual quality-control step across the processed reports. This final review focused on identifying residual translation errors, typographical mistakes, formatting inconsistencies, and garbled text. The resulting slide-level WSI--report pairs were used as the fine-grained multimodal component of TUM-Uteria.

\begin{figure*}[th]
  \centering
\includegraphics[width=1\textwidth]{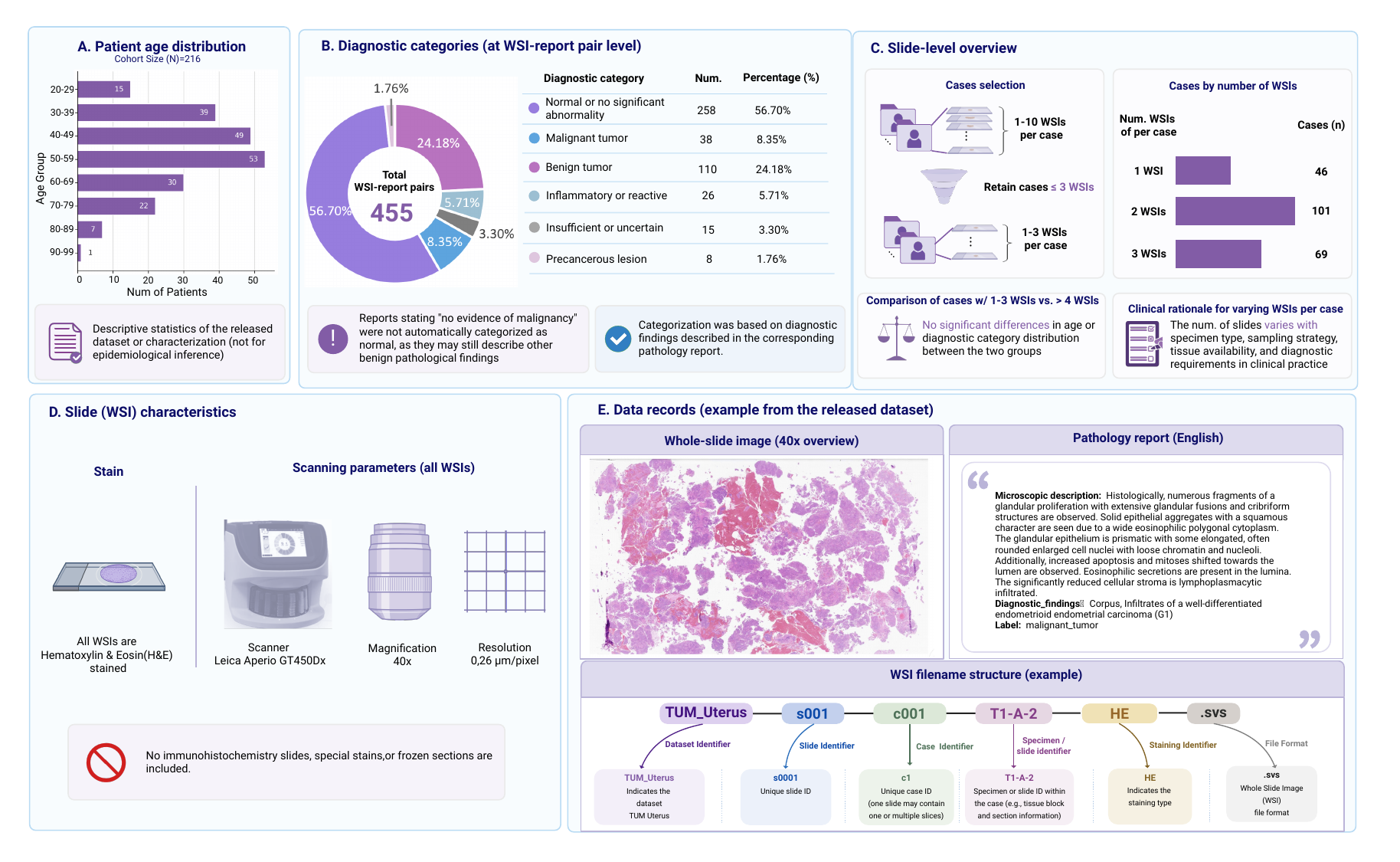}
\caption{
\textbf{Overview of the TUM-Uteria dataset.}
(A) Patient age distribution.
(B) Distribution of slide-level diagnostic categories.
(C) Overview of the number of WSIs per case.
(D) Characteristics of the released WSIs and scanning parameters.
(E) Example data records included in the released dataset, including WSIs, report-derived textual descriptions, diagnostic categories, and case-level identifiers.
}
 \label{fig2}
\end{figure*}
\section{Dataset characteristics}

We summarized the cohort-level, slide-level, and report-level characteristics of TUM-Uteria to provide an overview of the released dataset. These summaries are intended for dataset characterization and should not be interpreted as epidemiological evidence. TUM-Uteria includes \textbf{216 clinical cases} from female patients, with an age range of \textbf{23--92} years and a median age of \textbf{50} years. The age distribution is shown in Fig.~\ref{fig2}A, with most patients falling between \textbf{40} and \textbf{59} years. 

To facilitate downstream analysis and benchmarking, each WSI--report pair was assigned a slide-level diagnostic category derived from the diagnostic findings in the corresponding pathology report. The categories were grouped into six clinically meaningful classes: benign tumor (e.g. endometrial polyp and leiomyoma), inflammatory or reactive findings (e.g., endometritis and nonspecific reactive changes), insufficient or uncertain findings (e.g., scant, non-diagnostic, or indeterminate tissue samples), malignant tumor (e.g., endometrial carcinoma and uterine sarcoma), normal or no significant abnormality (e.g., proliferative, secretory, or atrophic endometrium without pathological findings), and precancerous lesion (e.g., atypical endometrial hyperplasia or endometrial intraepithelial neoplasia [EIN]). The numbers of WSI--report pairs in these categories were \textbf{110 (24.18\%)}, \textbf{26 (5.71\%)}, \textbf{15 (3.30\%)}, \textbf{38 (8.35\%)}, \textbf{258 (56.70\%)}, and \textbf{8 (1.76\%)}, respectively. The "normal or no significant abnormality" category represented specimens with evaluable tissue showing physiological, age-related, or otherwise non-pathological histological findings, rather than specimens without sufficient tissue for assessment. This distribution reflects the routine clinical composition of uterine pathology cases rather than a dataset selectively enriched for malignant or rare diseases.

Importantly, reports stating "no evidence of malignancy" were not automatically categorized as normal or healthy tissue, since such reports may still describe inflammatory, reactive, hemorrhagic, or other benign pathological findings. Cases were categorized as normal or no significant abnormality only when no clinically relevant pathological alteration was described in the report.

In the initial candidate cohort, cases contained between \textbf{1} and \textbf{10} WSIs. To reduce the ambiguity of slide--report alignment, we retained only cases with no more than \textbf{3} WSIs in the released slide-level paired dataset. We further compared cases with \textbf{1--3} WSIs and those with \textbf{4 or more} WSIs in terms of age and diagnostic category, and observed no significant differences between the two groups. The numbers of cases with 1, 2, and 3 WSIs were \textbf{46}, \textbf{101}, and \textbf{69}, respectively. This variation reflects routine diagnostic practice, where the number of slides depends on specimen type, sampling strategy, tissue availability, and diagnostic requirements. The distribution of WSIs per case is shown in Fig. \ref{fig2}.

All released WSIs are hematoxylin and eosin (H\&E)-stained slides. No immunohistochemistry slides, special stains, or frozen sections are included in the released dataset.


\section{Data Records}

The Uteria dataset is publicly available at [DATA REPOSITORY LINK]. The dataset contains whole-slide images and corresponding textual descriptions derived from pathology reports.

The dataset includes the following components:

\begin{itemize}

\item \textbf{Whole-slide images} stored in SVS format
\item \textbf{Slide-level textual descriptions (report)} extracted from pathology reports
\item \textbf{Slide-level diagnostic category} derived from the corresponding pathology report.
\item \textbf{Case-level identifiers (ID)} for grouping slides from the same patient and enabling case-level data partitioning.

\end{itemize}

\section{Technical Validation}

Several validation procedures were applied to ensure the reliability, privacy protection, and usability of the TUM-Uteria dataset. These procedures covered report anonymization, bilingual report quality, slide--report pairing, file integrity, and whole-slide image quality.

First, all diagnostic reports underwent a multi-step anonymization validation procedure. After the initial anonymization using a locally deployed offline Qwen-30B model, a second Qwen-30B-based screening step was performed to identify any residual patient-identifiable information. The processed reports were then manually reviewed by three pathology AI researchers to detect residual privacy-related content, incomplete de-identification, formatting artifacts, or other irregularities. Only reports that passed this anonymization validation were included in the final dataset.

Second, the bilingual report content was manually checked to ensure usability in both German and English. During the report processing and slide-level extraction steps, translation-related typographical errors, formatting inconsistencies, and garbled text were corrected when identified. After completion of the main processing pipeline, three pathology AI researchers performed an additional manual quality-control review across the processed reports to further identify residual translation errors, typographical mistakes, and formatting artifacts.

Third, the reliability of slide--report pairing was validated through expert review. For reports with explicit slide, block, or specimen identifiers, the extracted slide-level descriptions were checked to confirm that the corresponding WSI--report pairs were consistent with the original case-level diagnostic reports. For cases without clear slide-level identifiers, manual assessment was performed based on report structure, specimen information, slide metadata, and diagnostic context. In total, \textbf{71 cases} required manual judgment to determine the appropriate correspondence between WSIs and textual descriptions. These cases were specifically reviewed to establish reliable slide-level associations.

Fourth, all automatically extracted slide-level pairs that did not require manual assignment were further reviewed by two pathology experts to ensure consistency between the textual descriptions and the corresponding WSIs. This step was designed to verify that the derived slide-level annotations preserved the diagnostic meaning of the original case-level reports while enabling fine-grained visual--text correspondence.

Finally, automated scripts were used to verify file integrity, metadata completeness, and consistency between case identifiers, slide identifiers, WSI files, and associated report files. Whole-slide images were also visually inspected to identify severe scanning errors, major staining artifacts, blurred tissue regions, or other quality issues that could affect downstream computational analysis. Slides with significant quality issues were excluded from the dataset.

\section{Usage Notes}

The Uteria dataset can support a variety of research tasks in computational pathology, including:

\begin{itemize}

\item Whole-slide image analysis
\item Multimodal pathology AI
\item Pathology report generation
\item Visual–language learning for medical images

\end{itemize}

Researchers may tessellate WSIs into smaller patches for patch-level learning tasks or directly use the paired image–text data for training multimodal models. The dataset can also be used to evaluate the performance of visual–language models in pathology.

The dataset is distributed through the Hugging Face gated-access system. Users are required to create a Hugging Face account and agree to the Data Use Agreement prior to downloading the dataset. Access is granted automatically and does not require manual approval from the dataset maintainers. The access process is identical for all users, including peer reviewers. Reviewer identities are not disclosed to the authors through the registration process, and access is provided immediately after registration and agreement to the Data Use Agreement (DUA).

\section{Data Availability}

The Uteria dataset is available at \href{https://huggingface.co/datasets/Zhengyang-TUM/TUM_UterusReport}{HuggingFace: TUM\_UterusReport}.



\section{Author Contributions}

H.L. and J.L. conceived the dataset and designed the study. A.U., J.H., Z.X., A.K., O.T., C.G., F.G., R.N., X.M. and R.Y. collected, processed and curated the data. H.C. and S.K.Z. contributed methodological insights. N.N., C.M. and P.S. supervised the project. All authors contributed to manuscript preparation.

\section{Competing Interests}

The authors declare no competing interests.

\section{Acknowledgements}

We thank the pathology department staff for their assistance with data collection and slide digitization.


\section{Ethics statement}

This study used anonymized, retrospective pathology data, and
was approved by the Ethics Committee of the Technical University of Munich, in accordance with relevant ethical guidelines (2026-84-S-NP).

\printbibliography

\end{document}